\documentclass{article}
\usepackage[preprint]{colm2026_conference}

\usepackage{microtype}
\usepackage{hyperref}
\usepackage{url}
\usepackage{booktabs}
\usepackage{newpxtext} 
\usepackage{amsmath,amssymb,amsthm}
\usepackage{algorithm}
\usepackage{algorithmic}
\usepackage{lineno}

\definecolor{darkblue}{rgb}{0,0,0.5}
\hypersetup{colorlinks=true,citecolor=darkblue,linkcolor=darkblue,urlcolor=darkblue}

\theoremstyle{remark}

\theoremstyle{definition}

\title{HDPO: Hybrid Distillation Policy Optimization via Privileged Self-Distillation}

\author{Ken Ding \\
NVIDIA \\
\texttt{kennethd@nvidia.com}
}

\begin{document}

\ifcolmsubmission
\linenumbers
\fi

\maketitle

\begin{abstract}
Large language models trained with reinforcement learning (RL) for mathematical reasoning face a fundamental challenge: on problems the model cannot solve at all---``cliff'' prompts---the RL gradient vanishes entirely, preventing any learning signal from reaching these failure modes. We introduce Hybrid Distillation Policy Optimization (HDPO), which augments standard RL with privileged self-distillation targeting cliff prompts. On each training step, HDPO identifies prompts where all rollouts fail, generates privileged rollouts by providing the model with ground-truth information, filters for correct solutions, and distills the teacher's token-level distribution into the student. Because teacher and student share the same weights---differing only in their input---the realizability gap is provably bounded, unlike cross-model distillation. We prove that $R{=}1$ filtered privileged generation recovers the optimal KL-regularized RL policy in the hard-threshold limit. Experiments on OpenMathInstruct-2 with Qwen2.5-Math-1.5B-Instruct show that HDPO consistently improves coverage metrics (\textit{pass@4} by $+0.8$--$1.1\%$, \textit{pass@8} by $+0.4$--$1.7\%$) while maintaining greedy accuracy, with the distillation weight $\lambda$ providing direct control over the exploration--exploitation tradeoff.
\end{abstract}

\section{Introduction}
\label{sec:intro}

Recent advances in reinforcement learning from verifiable rewards (RLVR)
have enabled large language models to develop sophisticated mathematical
reasoning capabilities \citep{Shao2024,DeepSeekAI2025}. Among RL algorithms for language models,
Group Relative Policy Optimization (GRPO)
\citep{Shao2024} has emerged as a practical and
effective approach, eliminating the need for a critic network by
normalizing rewards within each group of rollouts to compute advantages.

However, GRPO and related policy gradient methods share a fundamental
limitation: they can only learn from problems where at least some
rollouts succeed. When all rollouts for a prompt receive zero reward ---
a condition commonly called a "cliff" \citep{ZhangX2025} --- the advantage estimates are identical
for all trajectories and the policy gradient vanishes. These
zero-gradient prompts are precisely the ones where learning is most
needed, yet they receive no training signal. Le et al.
\citep{Le2025} characterize this as the zero-variance
prompt problem and show that, on their benchmark, a significant fraction
of prompts falls in this regime---the exact proportion depends on the
dataset and model, but the phenomenon creates a persistent learning dead
zone at the frontier of the model's capability.

We take a different approach inspired by the principle of learning using
privileged information\citep{Vapnik2009}. In our framework,
Hybrid Distillation Policy Optimization (HDPO), the model acts as both
teacher and student. As the teacher, it receives the problem along with
ground-truth information and generates rollouts from this privileged
context. As the student, it receives only the original problem. Because
both roles use the same model weights, the gap between their
distributions is provably bounded (Proposition 1), unlike cross-model distillation where the gap depends on architectural differences between teacher and student.

HDPO operates as follows: on each training step, after standard GRPO
updates, we identify cliff prompts where all rollouts scored zero. For
these prompts, the model generates privileged rollouts conditioned on
ground truth, filters for correct ones (\(R\  = \ 1\)), and distills the
teacher's token-level distribution into the student via JSD. The
\(R\  = \ 1\) filter is not a heuristic --- we prove it implements
rejection sampling from the KL-regularized RL optimal policy
(Proposition 2).

Our contributions are as follows:

(1) We introduce HDPO, a hybrid training objective that combines RL with
privileged self-distillation targeting cliff prompts where the RL
gradient vanishes.

(2) We prove that same-model privileged distillation achieves a strictly
tighter realizability gap than cross-model distillation --- the gap
depends only on the information content of the ground truth, eliminating
the model-mismatch term that cross-model distillation cannot avoid
(Proposition 1).

(3) We prove that \(R\  = \ 1\) filtered privileged generation recovers
the optimal KL-regularized RL policy (Proposition 2), providing
theoretical justification for the teacher construction.

(4) We demonstrate on OpenMathInstruct-2
\citep{Toshniwal2024} that HDPO improves coverage
(\(pass@4\), \(pass@8\)) while maintaining greedy accuracy (\(pass@1\)),
with the distillation weight \(\lambda\) providing explicit control over
the exploration--exploitation tradeoff.

\section{Background}
\label{sec:background}\label{background}

\subsection{Reinforcement Learning for Language
Models}

Reinforcement learning-based training of language models formulates text
generation as a sequential decision problem where the policy
\(\pi_{\theta}\) generates tokens autoregressively and receives a scalar
reward upon completion. This framework, popularized by RLHF
\citep{Ouyang2022} for instruction following, has been
adapted to mathematical reasoning as reinforcement learning from
verifiable rewards (RLVR), where the reward is typically binary:
\(r(x,y) = 1\) if the solution y is correct and 0 otherwise
\citep{DeepSeekAI2025}. This binary, outcome-level reward
makes credit assignment particularly challenging, as the model must
determine which tokens contributed to a correct or incorrect answer.

\subsection{Group Relative Policy
Optimization}

GRPO \citep{Shao2024} generates \(G\) rollouts
\(\left\{ y_{1},\ \ldots,\ y_{G} \right\}\) for each prompt \(x\) and
computes advantages by normalizing rewards within each group:
\(\widehat{A_{i}}\  = \ \frac{r_{i}\  - \ \mu}{\sigma}\). This
eliminates the need for a separate critic network, reducing memory
requirements and avoiding the instability of value function estimation.

GRPO applies PPO-style ratio clipping \citep{Schulman2017} to
stabilize updates:
\(L_{GRPO}\  = \  - E\left\lbrack \min\left( \rho_{t}A,\ clip\left( \rho_{t},\ 1 - \varepsilon,\ 1 + \varepsilon \right)A \right) \right\rbrack\),
where
\(\rho_{t}\  = \ \frac{\pi_{\theta}\left( a_{t}|s_{t} \right)}{\pi_{old}\left( a_{t}|s_{t} \right)}\)
is the importance ratio. Training alternates between generating rollouts
from the current policy and performing gradient updates on the clipped
objective.

\subsection{The Cliff Problem}

Under binary reward with G rollouts, the advantage for a prompt x takes
one of three forms: (i) all rollouts succeed, giving zero advantages ---
the model has already learned this prompt; (ii) mixed results, giving
positive advantages for successes and negative for failures --- the
standard learning regime; or (iii) all rollouts fail, giving zero
advantages --- the model receives no gradient despite urgently needing
signal.

Case (iii) --- commonly called a ``cliff'' \citep{ZhangX2025} --- is the critical failure
mode: the hardest problems, which represent the frontier of the model's
capability, receive essentially no gradient
\citep{Le2025}. The model can only learn from prompts
at intermediate difficulty --- hard enough that some rollouts fail
(providing contrast for advantage estimation) but easy enough that some
succeed. The cliff boundary can only advance indirectly, as learning on nearby prompts
must transfer to cliff prompts through weight sharing alone --- there is no
direct gradient signal on the cliffs themselves.

\subsection{Approaches to the Cliff
Problem}

A growing body of work has identified the cliff problem as a critical
bottleneck in RL for reasoning and proposed diverse strategies to
address it. These approaches introduce substantial additional complexity
--- new hyperparameters, auxiliary models, replay infrastructure, or
modifications to the training loop --- to work around the zero-gradient
problem. By contrast, HDPO requires only a single additional forward
pass with ground truth appended to the prompt and a standard JSD loss:
no curriculum scheduler, no replay buffer, no process reward model, no
scaffolding heuristics. We organize the existing approaches into several
categories to illustrate this complexity.

\textbf{Curriculum and filtering}. VCRL \citep{Jiang2025}
observes that group reward variance is a natural proxy for prompt
difficulty and dynamically schedules sampling to focus on the
high-variance frontier. DAPO \citep{Yu2025dapo} filters
out zero-variance prompts entirely, oversampling until informative prompts
are found. While
effective, these methods avoid cliff prompts rather than learning from
them --- the hardest problems are simply skipped,
leaving the cliff boundary to advance only through indirect transfer
from nearby solved prompts.

\textbf{Scaffolded and augmented generation}. Scaf-GRPO
\citep{ZhangX2025} diagnoses learning stagnation on cliff
prompts and injects tiered in-prompt hints --- from abstract concepts to
concrete steps --- that fade as the model improves. HINT
\citep{Wang2025hint} provides targeted guidance to help
ineffective rollouts navigate toward correct solutions. EvoCoT
\citep{Liu2025evocot} constrains the exploration space by
self-generating and verifying chain-of-thought trajectories, then
gradually expands by shortening CoT steps. These methods require
designing or learning a hint generation strategy, deciding when to
inject and fade hints, and managing the distributional gap between
scaffolded and unscaffolded generations --- a non-trivial engineering
and research effort per domain.

\textbf{Experience replay}. Retrospective Replay
\citep{Dou2025} stores early exploratory trajectories
and revisits them later when the model has grown capable. RLEP
\citep{ZhangH2025rlep} collects verified correct trajectories
and blends them into mini-batches. Replay methods introduce a buffer
management system with its own hyperparameters (buffer size, sampling
strategy, staleness thresholds) and face a fundamental tension: replayed
trajectories are off-policy and may become increasingly stale as
training progresses, potentially teaching the model to imitate outdated
reasoning patterns.

\textbf{Advantage shaping and process rewards}. Le et al.
\citep{Le2025} tackle zero-variance prompts by scaling
each token's gradient in proportion to its entropy, reclaiming learning
signal from otherwise wasted prompts. PRIME
\citep{Cui2025prime} enables online process reward model
updates using only policy rollouts and outcome labels. Lightman et al.
\citep{Lightman2024} demonstrate that step-level
verification rewards substantially improve reasoning. However, process
reward approaches require training and maintaining a separate reward
model --- itself a significant undertaking --- and shaped rewards may
not perfectly align with the outcome objective.

\textbf{Interleaved RL and distillation}. ReLIFT
\citep{Ma2025} finds that RL excels on easier
questions while SFT is crucial for the hardest ones, and interleaves RL
with targeted online SFT on questions the current policy cannot solve.
ReLIFT is conceptually closest to HDPO but is considerably more complex:
it requires a two-phase training loop with switching logic, an external
source of high-quality solutions for the SFT phase, and careful tuning
of the interleaving schedule. HDPO achieves the same goal --- learning
signal on cliff prompts --- with a single unified objective, using the
model's own privileged rollouts rather than external solutions, and with
theoretical guarantees (bounded realizability gap, optimal target
characterization) that ReLIFT lacks.

In summary, existing approaches to the cliff problem introduce
substantial machinery --- curriculum schedulers, hint generators, replay
buffers, process reward models, or multi-phase training loops --- to
work around the zero-gradient problem. HDPO's mechanism is remarkably
simple by comparison: append ground truth to the prompt, generate,
filter for correctness, and distill via JSD. This simplicity is not a
limitation but a feature: it arises from the observation that the model
can solve cliff prompts when given privileged information --- the same
weights that fail unprompted succeed with ground-truth context, providing
a naturally bounded distillation target. HDPO is also
complementary to these approaches --- curriculum methods could
prioritize near-cliff prompts, and process rewards could densify signal
on non-cliff prompts --- but it achieves its core effect with minimal
additional complexity.

\subsection{Knowledge Distillation for
Reasoning}

Knowledge distillation \citep{Hinton2015} transfers
knowledge from a teacher to a student by minimizing the divergence
between their output distributions. For autoregressive language models,
a key challenge is distribution mismatch: the student is trained on
teacher-generated or fixed sequences, but at inference must generate
from its own distribution. Agarwal et al.
\citep{Agarwal2024gkd} address this with Generalized Knowledge
Distillation (GKD), which trains the student on its own generated
outputs while receiving teacher feedback. GKD offers flexibility in
divergence choice (forward KL, reverse KL, JSD).

\textbf{Self-distillation for reasoning}. Zhao et al.
\citep{Zhao2026} propose on-policy self-distillation
(OPSD), where a single LLM serves as both teacher and student under
different informational contexts --- the teacher conditions on
privileged information such as verified reasoning traces or ground-truth
solutions, while the student sees only the problem. Training minimizes
per-token divergence over the student's own on-policy rollouts. H\"{u}botter
et al. \citep{Hubotter2026} introduce Self-Distillation
Policy Optimization (SDPO), which converts rich textual feedback
(runtime errors, judge evaluations) into a dense learning signal by
distilling the model's feedback-conditioned predictions back into the
unconditional policy, addressing the credit-assignment bottleneck of
scalar-only reward.

\textbf{Unified KD and RL frameworks.} Several works jointly optimize
distillation and RL objectives. KDRL \citep{Xu2025kdrl}
simultaneously minimizes reverse KL divergence between student and
teacher while maximizing expected reward, finding that the combination
outperforms either objective alone. RLAD
\citep{ZhangZ2026rlad} replaces the standard importance ratio
with a geometric mixture of the old policy and a teacher model,
embedding the teacher directly into the RL policy update. G-OPD
\citep{Yang2026gopd} shows theoretically that standard
on-policy distillation is a special case of dense KL-constrained RL and
introduces reward extrapolation to enable students to surpass teacher
performance.

\textbf{Diversity preservation}. Li et al.
\citep{Li2025dph} identify the diversity collapse
problem: RLVR fine-tuning improves pass@1 but degrades pass@k as the
policy concentrates around a single mode. DPH-RL replaces mode-seeking
divergences (reverse KL) with mass-covering alternatives (forward KL,
JSD) that continuously reference the initial policy to maintain broad
solution coverage. This insight directly motivates HDPO's use of JSD for
the distillation loss: JSD provides mode-covering signal that expands
the student's support without collapsing to a single mode.

\textbf{Privileged information}. The concept of learning using
privileged information (LUPI) \citep{Vapnik2009} provides a
theoretical framework for training with auxiliary information available
only at training time. Penaloza et al. \citep{Penaloza2026}
introduce pi-Distill, a joint teacher-student objective for privileged
information distillation. HDPO instantiates the LUPI framework
specifically for RL: ground truth serves as privileged information, the
model itself serves as teacher, JSD \citep{Lin1991}
provides the distillation mechanism, and distillation targets only cliff
prompts where the RL gradient vanishes.

\section{Method: Hybrid Distillation Policy
Optimization}

\subsection{Motivation}\label{motivation}

The central insight of HDPO is that for cliff prompts, we can construct
a teacher signal without any external model or human annotation --- by
giving the model itself access to the answer. When conditioned on
ground-truth information (e.g., the solution to a math problem), even a
small model can generate correct reasoning traces with high probability.
The key properties of this construction are: (i) the teacher and student
share the same weights, differing only in their input context; (ii) the
\(R = 1\) filter selects only correct teacher trajectories; and (iii)
distillation targets only cliff prompts where the RL gradient is zero.

Property (i) is critical: because teacher and student are the same
model, the realizability gap --- the KL divergence between their output
distributions --- is bounded by a function of the privileged information
alone (Proposition 1). This is strictly tighter than the bound for any
cross-model teacher, which incurs an additional model-mismatch term that
same-model distillation eliminates entirely. Property (ii) ensures we
distill from the optimal target, not just any correct trajectory.
Property (iii) restricts distillation to cliff prompts, where the RL
gradient is zero and thus cannot provide signal on its own.

\subsection{Formal Definition}

\paragraph{Formal Definition.} The HDPO training objective combines
\(\mathcal{L}_{GRPO}\) with a \(\mathcal{L}_{JSD}\) distillation term on
cliff prompts:

\[\mathcal{L}_{HDPO}(\theta)\  = \ \mathcal{L}_{GRPO}(\theta)\  + \ \lambda\  \cdot \ \mathcal{L}_{JSD}(\theta)\]

where \(\mathcal{L}_{JSD}\) is the token-averaged JSD over filtered
teacher trajectories on cliff prompts:

\[\mathcal{L}_{JSD}(\theta)\  = \ \frac{1}{N_{tok}}\sum_{(x,\ y)\mathcal{\in T}}^{}{\sum_{t\  = \ 1}^{|y|}{{JSD}_{k}\left( \pi_{T}\left( \  \cdot \  \mid \ y_{< \ t} \right)\  \parallel \ \pi_{\theta}\left( \  \cdot \  \mid \ y_{< \ t} \right) \right)}}\]

The distillation set \(\mathcal{T}\) is constructed via two levels of
filtering. First, identify cliff prompts --- prompts where all \(K\)
standard rollouts failed:

\[\mathcal{C} = \{x \in \mathcal{B} : \textstyle\sum_k R(x, y^{(k)}) = 0\}\]

Second, for each \(x\  \in \ \mathcal{C}\), generate privileged rollouts
\(\bar{y}_{j}\  \sim \ \pi_{\theta}( \cdot \ |\ x\  \oplus \ y*)_{x}\) by
injecting ground truth \(y^{*}\) into the prompt, and retain only
correct trajectories:

\[\mathcal{T} = \{(x,\ \bar{y})\ :\ x \in \mathcal{C},\ \bar{y} \sim \pi_{\theta}(\cdot \mid x \oplus y^*),\ R(x,\ \bar{y}) = 1\}\]

That is, \(\mathcal{T}\) contains (prompt, trajectory) pairs where (a)
the prompt is a cliff --- all standard rollouts scored zero --- and (b)
the privileged model, conditioned on ground truth, generated a correct
solution.
\(N_{tok}\  = \ \sum_{\left( x,\ \overline{y} \right)\mathcal{\in T}}^{}\left| \overline{y} \right|\)
is the total distillation token count (computed globally across all
data-parallel ranks). \(\pi_{T}\) and \(\pi_{\theta}\) share the same
weights; they differ only in their input (privileged vs. unprivileged).
In practice, \({JSD}_{k}\) is approximated using the
teacher\textquotesingle s top-\(k\) (\(k\  = \ 64\)) logits,
renormalized. The tail correction for student mass outside the top-\(k\)
support is exact: \(P_{rest}\  \cdot \ ln\ 2\).

\subsection{Algorithm}\label{algorithm}

\begin{algorithm}[t]
\caption{HDPO Training}
\label{alg:hdpo}
\begin{algorithmic}[1]
\REQUIRE Policy $\pi_\theta$, prompt set $\mathcal{X}$, ground truth $\{y^*\}$,
         reward $R$, learning rate $\alpha$, distillation weight $\lambda$,
         rollouts per prompt $K$
\FOR{step $= 1, \ldots, N$}
  \STATE \textit{// Standard GRPO}
  \STATE Sample prompt batch $B \subset \mathcal{X}$
  \FORALL{$x \in B$}
    \STATE Generate $K$ rollouts $y^{(k)} \sim \pi_\theta(\cdot \mid x)$
  \ENDFOR
  \STATE Score: $\hat{A}_i = (r_i - \mu)/\sigma$
  \STATE Compute $\mathcal{L}_\mathrm{GRPO}$ via clipped policy gradient with leave-one-out advantages
  \STATE \textit{// Privileged self-distillation on cliff prompts}
  \STATE Identify cliffs: $\mathcal{C} = \{x \in B : \textstyle\sum_k r^{(k)} = 0\}$
  \FORALL{$x \in \mathcal{C}$}
    \STATE Generate $\bar{y}^{(j)} \sim \pi_\theta(\cdot \mid x \oplus y^*)$
  \ENDFOR
  \STATE Filter: $\mathcal{T} = \{(x, \bar{y}) : R(x, \bar{y}) = 1\}$
  \STATE Compute $\mathcal{L}_\mathrm{JSD} = \frac{1}{N_\mathrm{tok}}\sum_{(x,\bar{y})\in\mathcal{T}}\sum_{t=1}^{|\bar{y}|}\mathrm{JSD}_k(\pi_T(\cdot|\bar{y}_{<t})\,\|\,\pi_\theta(\cdot|\bar{y}_{<t}))$
  \STATE \textit{// Update}
  \STATE $\theta \leftarrow \theta - \alpha\nabla_\theta\bigl(\mathcal{L}_\mathrm{GRPO}(\theta) + \lambda\cdot\mathcal{L}_\mathrm{JSD}(\theta)\bigr)$
\ENDFOR
\end{algorithmic}
\end{algorithm}

\subsection{Theoretical Analysis}

We argue that HDPO's effectiveness rests on two properties: on the
correct teacher trajectories, the distance between the student and
teacher distributions is bounded (unlike cross-model distillation), and
the \(R\  = \ 1\) filter ensures the teacher distribution the student
moves toward corresponds to the KL-regularized RL optimal policy
(Proposition 2).
Moreover,
this bound is strictly tighter than what any cross-model teacher can
achieve: because teacher and student share the same function, the only
source of distributional divergence is the privileged information
itself. A cross-model teacher introduces an additional model-mismatch
term that same-model distillation eliminates entirely (Proposition 1).

\textbf{Proposition 1} (Realizability Gap Comparison)\label{thm:realizability}. \emph{Let} \(P_{T}\)
\emph{and} \(P_{S}\) \emph{denote the per-position output distributions
of a language model} \(\theta\) \emph{when prompted with and without
ground truth} \(g\)\emph{, respectively.}

\[D_{KL}(P_{T} \parallel P_{S}) \leq \frac{L_{\theta}^{2} \cdot \Delta(g)^{2}}{2}\]

\textbf{(i) Same-model bound.} \emph{where} \(L_{\theta}\) \emph{is the
local Lipschitz constant of the model's logit function on bounded inputs
and} \(\Delta(g)\) \emph{is the input-space distance attributable to}
\(g\)\emph{.}

\emph{\textbf{(ii) Cross-model comparison.} For cross-model distillation
with a separate teacher model} \(\phi\) \emph{receiving the same
privileged input} \(c_{T}\)\emph{, the realizability gap satisfies:}

\[D_{KL}(P_{\phi}( \cdot |c_{T}) \parallel P_{\theta}( \cdot |c_{S})) \leq \frac{\left( L_{\theta} \cdot \Delta(g) + \left\| f_{\phi}\left( c_{T} \right) - f_{\theta}\left( c_{T} \right) \right\|_{\infty} \right)^{2}}{2}\]

\emph{The cross-model bound contains an additive model-mismatch term}
\(\left\| f_{\phi}\left( c_{T} \right) - f_{\theta}\left( c_{T} \right) \right\|_{\infty}\)
\emph{that is absent from the same-model bound. This term vanishes if
and only if the models produce identical logits on the privileged
input.}

\paragraph{Proof sketch.} Both bounds follow from Lemma 1 (Appendix A),
which shows
\(D_{KL} \leq \frac{{\left\| \delta \right\|_{\infty}}^{2}}{2}\) for any
logit perturbation \(\delta\). For (i): since teacher and student
evaluate the same function \(f_{\theta}\) on different inputs, the logit
difference is bounded by \(L_{\theta} \cdot \Delta(g)\) via Lipschitz
continuity. For (ii): the logit difference decomposes via the triangle
inequality into a model-mismatch term plus the same input-perturbation
term from (i) (Appendix A).

\paragraph{Interpretation.} The proposition provides a direct comparison
between same-model and cross-model distillation. Both bounds share the
same \(\Delta(g)\) term --- the cost of incorporating privileged
information --- but the cross-model bound pays an additional price for
model mismatch. With a \textbf{drifting teacher} (where the teacher
shares the student's current weights), the model-mismatch term is
identically zero throughout training. With a \textbf{frozen teacher}
(initialized at the start of training), a non-zero model-mismatch term
appears, yielding looser guarantees that approach the cross-model regime.

\paragraph{Relation to prior work.} The bound follows from standard
Lipschitz continuity and logit perturbation arguments. Lemma~1
(Appendix~A) is a known bound on KL divergence under softmax logit
perturbation. The contribution of Proposition~1 is not the bound itself
but the comparison: it makes explicit that same-model privileged
distillation eliminates the model-mismatch term that cross-model
distillation cannot avoid.

\textbf{Proposition 2}\label{prop:filtering} ($R{=}1$ Filtering Yields the RL-Optimal
Policy). \emph{Consider the KL-regularized RL objective:}

\[\max_{\pi}\ \mathbb{E}_{\tau\sim\pi}\lbrack R(\tau)\rbrack\  - \ \beta\  \cdot \ KL(\pi\  \parallel \ \pi_{ref})\]

\emph{For binary reward} \(R \in \{ 0,1\}\)\emph{, the unique optimizer
is} \(\pi^{*}(x) \propto \pi_{ref}(x)\) \(exp(R(x)/\beta)\)\emph{. In
the limit} \(\beta \rightarrow 0^+\)\emph{,} \(\pi^{*}\) \emph{converges
to} \(\pi_{ref}( \cdot |R = 1)\) \emph{--- the reference policy
conditioned on correctness --- and} \(R = 1\) \emph{rejection sampling
from} \(\pi_{ref}\) \emph{recovers it exactly.}

\paragraph{Proof sketch.} The unique solution takes the form~\citep{Rafailov2023}:

\[\pi^{*}(\tau) = \ \pi_{ref}(\tau) \cdot \frac{\exp\left( \frac{R(\tau)}{\beta} \right)}{Z(\beta)}\]

where \(Z(\beta)\) is the partition function. For binary reward
\(R(\tau) \in \{ 0,1\}\), this reweights correct trajectories by
\(exp(1/\beta)\) relative to incorrect ones. In the hard-threshold limit
\(\beta \rightarrow 0^+\), incorrect trajectories receive zero weight and
the optimal policy concentrates entirely on correct trajectories
(Appendix B):

\[\pi^{*}(\tau)\  \rightarrow \ \pi_{ref}(\tau\ |\ R(\tau)\  = \ 1)\  = \ \frac{\pi_{ref}(\tau)\  \cdot \ 1\lbrack R(\tau)\  = \ 1\rbrack}{P_{\pi_{ref}}(R\  = \ 1)}\]

\paragraph{Connection to HDPO.} The KL-regularized objective above is the
standard RL formulation: maximize reward while staying close to a
reference \(\pi_{ref}\). Proposition 2 shows that in the
\(\beta\  \rightarrow \ 0\) limit with binary reward, the optimal policy
is \(\pi_{ref}( \cdot \ |\ R\  = \ 1)\) --- the reference distribution
conditioned on correctness --- and that \(R = 1\) rejection sampling
recovers it exactly. On non-cliff prompts, GRPO\textquotesingle s policy
gradient optimizes this objective directly and provides learning signal.
On cliff prompts, however, all \(K\) rollouts score zero:
\(P_{\pi\_\theta( \cdot |prompt)}(R\  = \ 1)\  \approx \ 0\). The
reference has negligible support on correct trajectories, so
\(\pi_{ref}( \cdot \ |\ R\  = \ 1)\) is inaccessible --- rejection
sampling produces no accepted samples, and the RL gradient vanishes.

HDPO resolves this by replacing \(\pi_{ref}\) with a proxy whose optimal
solution has non-degenerate support on correct trajectories. Injecting
ground truth \(g\) into the prompt yields
\(\pi_{\theta}( \cdot \ |\ prompt,\ g)\), which shares the
student\textquotesingle s weights but has substantially higher
\(P(R\  = \ 1)\). The KL-regularized objective with this proxy
reference,

\[\max\ _{\pi}\mathbb{\ E\lbrack}R(\tau)\rbrack\  - \ \beta\  \cdot \ KL(\pi\  \parallel \ \pi_{\theta}( \cdot \ |\ prompt,\ g))\]

has a well-defined optimal solution
\(\pi_{\theta}( \cdot \ |\ prompt,\ g,\ R\  = \ 1)\) with non-empty
support on correct trajectories, precisely where the original
objective\textquotesingle s solution
\(\pi_{\theta}( \cdot \ |\ prompt,\ R\  = \ 1)\) had none. Ground truth
injection expands the set of trajectories \(\tau\) where
\(R(\tau)\  = \ 1\), transforming an objective with a degenerate
solution into one with a substantive target. \(R = 1\) filtering on this
proxy has a well-defined optimal solution (Proposition 2). HDPO
approximates it by generating from \(\pi_\theta(\cdot \mid x, y^*)\) and
retaining only R=1 completions, yielding a finite-sample estimate of
\(\pi_\theta(\cdot \mid x, y^*, R{=}1)\). This departs from the theoretical target
in two ways: the proxy reference \(\pi_\mathrm{priv}\) replaces \(\pi_\mathrm{ref}\) (the gap
bounded by Proposition 1), and the privileged pass rate can be substantially
below 1 on hard problems, introducing sampling noise. Proposition 1 then
bounds the gap between this proxy target and the ideal target
\(\pi_{\theta}( \cdot \ |\ prompt,\ R\  = \ 1)\): because teacher and
student share weights, the only source of divergence is the privileged
information itself, with the model-mismatch term identically zero for a
drifting teacher.

$\pi$-Distill \citep{Penaloza2026} independently arrives at the same objective
structure: its teacher objective
\(J_{Teacher}\mathbb{\  = \ E\lbrack}R\rbrack\  - \ \beta\  \cdot \ D_{KL}(\pi_{T}( \cdot |x,\ I)\  \parallel \ sg\lbrack\pi_{S}( \cdot |x)\rbrack)\)
takes exactly this form with \(\pi_{ref}\) = sg{[}\(\pi_{S}\){]}. Where
$\pi$-Distill optimizes this objective via gradient ascent at finite
\(\beta\) across all prompts, HDPO implements the
\(\beta\  \rightarrow \ 0\) limit mechanically via hard R=1 filtering,
activating only on cliff prompts where gradient-based optimization of
the objective fails.

\section{Experiments}
\label{sec:experiments}\label{experiments}

\subsection{Experimental Setup}

We evaluate HDPO using OpenMathInstruct-2
\citep{Toshniwal2024}, a large-scale dataset of mathematical
problems sourced from the MATH competition benchmark
\citep{Hendrycks2021} and GSM8K. We use
Qwen2.5-Math-1.5B-Instruct \citep{Yang2024qwen} as the base
model. The dataset is split into 95\% training and 5\% validation (seed
42), yielding 2048 validation problems evaluated with binary rewards
from hf\_math\_verify. The base RL algorithm is GRPO with G=16 rollouts
per prompt, leave-one-out advantages, and PPO-style ratio clipping
($\varepsilon{=}0.2$). We train for 2000 steps with AdamW
\citep{Loshchilov2019} at learning rate $1\mathrm{e}{-}6$ with linear warmup.
Generation uses vLLM \citep{Kwon2023} at temperature
1.0. Full experimental details are provided in Appendix C.

We compare four HDPO configurations against the GRPO baseline, varying
the teacher type (frozen at initialization vs. drifting with current
policy) and the distillation weight (\(\lambda\) $\in$ \{0.01, 0.1\}). All
HDPO variants use top-k=64 JSD with global token-count normalization. We
report best pass@1, pass@4, and pass@8 across training, evaluated every
10 steps with 2048 samples per evaluation.

\subsection{Results}\label{results}

\begin{table}[t]
\centering
\caption{Best pass@$k$ on OpenMathInstruct-2 validation (2048 samples,
Qwen2.5-Math-1.5B-Instruct) across 2000 training steps on 8$\times$H200 GPUs.
All HDPO variants use JSD with global token-count normalization.
Bold indicates best in column. Results on 8$\times$H100 GPUs are provided in Appendix~\ref{app:hardware} and show consistent trends with minor quantitative variation.}
\label{tab:results}
\vspace{2pt}
\begin{tabular}{lccc}
\toprule
\textbf{Method} & \textbf{pass@1} & \textbf{pass@4} & \textbf{pass@8} \\
\midrule
GRPO Baseline                & \textbf{0.6519} & 0.7749          & 0.8228 \\
HDPO (frozen, $\lambda$=0.01)   & \textbf{0.6519} & 0.7812          & 0.8218 \\
HDPO (frozen, $\lambda$=0.1)    & 0.6304          & 0.7812          & \textbf{0.8398} \\
HDPO (drifting, $\lambda$=0.01) & 0.6514          & \textbf{0.7861} & 0.8271 \\
HDPO (drifting, $\lambda$=0.1)  & 0.6294          & 0.7856          & 0.8364 \\
\bottomrule
\end{tabular}
\end{table}

The drifting teacher at \(\lambda\  = \ 0.01\) achieves the highest
pass@4 (0.7861, +1.1\% over baseline) and strong pass@8 (0.8271, +0.4\%) while essentially
maintaining pass@1 (0.6514 vs 0.6519). This is our \textbf{primary
result} --- HDPO broadens the model's coverage without sacrificing
greedy accuracy. We note that the magnitude of improvement at $\lambda{=}0.01$ varies across hardware (see Appendix~\ref{app:hardware}); the effect is consistent in direction but modest in size relative to run-to-run variance.

Increasing \(\lambda\) from 0.01 to 0.1 trades pass@1 for pass@8. Both
teacher types at \(\lambda\  = \ 0.1\) achieve \textasciitilde0.84
pass@8 (+1.4--1.7\% over baseline) but pass@1 drops by
\textasciitilde2.3--2.8\%. \(\lambda\) directly controls the
exploration--exploitation trade-off in the output distribution. The $\lambda{=}0.1$ improvements on pass@8 are the most robust finding, reproducing consistently across hardware configurations.

The drifting teacher advantage is most pronounced at low \(\lambda\);
at \(\lambda\  = \ 0.1\), the advantage narrows as the strong
distillation signal dominates.

\section{Discussion}
\label{sec:discussion}\label{discussion}

The results demonstrate that HDPO provides a consistent mechanism for
broadening the model's solution distribution on mathematical reasoning
tasks. At $\lambda{=}0.01$, HDPO improves pass@4 and pass@8 while essentially maintaining pass@1, though the magnitude of improvement is modest and varies across hardware (Appendix~\ref{app:hardware}). At $\lambda{=}0.1$, the coverage gains are larger and more robust, at the cost of pass@1. This is the core value proposition: by providing
learning signal on cliff prompts where GRPO cannot, HDPO expands the
model's coverage, with $\lambda$ controlling how aggressively it trades greedy accuracy for coverage.

The \(\lambda\) parameter provides explicit control over the
exploration--exploitation tradeoff. At \(\lambda\)=0.01, the
distillation signal is a gentle nudge that broadens coverage; at
\(\lambda\)=0.1, it becomes the dominant training signal on cliff
prompts, significantly improving pass@8 but at the cost of pass@1. At our model scale (1.5B), we observe a consistent tradeoff between
pass@1 and pass@$k$ as \(\lambda\) increases, though it remains an open
question whether this tradeoff persists at larger scales.

We hypothesize that this tradeoff reflects model capacity interacting
with mode-covering distillation. When the privileged model solves a
problem via multiple distinct strategies, the JSD loss trains the
student to place mass on all of them. For a small model, limited
capacity means these modes compete for the same parameters: the model
cannot maintain multiple fully coherent reasoning strategies
simultaneously. The result is a flatter distribution where no single
strategy dominates cleanly, degrading greedy accuracy (pass@1), while
the broader support means additional samples discover genuinely
different approaches (improving pass@k). We hypothesize that the ideal output distribution
is not uniform but concentrated: a single dominant mode that greedy
decoding reliably recovers, with smaller secondary modes accessible
through repeated sampling. Pure RL naturally produces this shape through
mode-seeking dynamics but cannot discover strategies where all rollouts
fail. This analysis motivates the expand-then-sharpen curriculum
described in Section 6: distillation first broadens strategy support on
cliff prompts, then RL restores a dominant mode while preserving
secondary strategies as a long tail.

The drifting teacher --- which shares the current policy's weights at
each step --- outperforms the frozen teacher (initialized at the start
of training) at low \(\lambda\). At \(\lambda = 0.1\), the advantage
narrows as the strong distillation signal dominates.

The frozen teacher at \(\lambda\)=0.1 achieves the single highest pass@8
(0.8398), suggesting that the initial model's distribution may be more
diverse --- not yet shaped by RL's mode-seeking dynamics --- at the cost
of a larger realizability gap. These results suggest a tradeoff between signal diversity and
realizability in teacher choice, though confirming this as a general
principle requires broader evaluation across scales and datasets.

\section{Limitations}
\label{sec:limitations}\label{limitations}

Our results are demonstrated on a single model scale (1.5B parameters)
and a single dataset (OpenMathInstruct-2, sourced from MATH
\citep{Hendrycks2021}). While the theoretical analysis (Proposition
1, Proposition 2) is scale-agnostic, the empirical magnitude of HDPO's
improvements may vary with model capacity, dataset difficulty, and
reward function design. Larger models may have fewer cliff prompts
(higher baseline success rates), potentially reducing HDPO's marginal
benefit, or may benefit more from the expanded coverage.

HDPO introduces computational overhead: generating and filtering
privileged rollouts, computing top-k teacher logits, and the additional
forward pass for the JSD loss. In our implementation, privileged
generation is performed by the same vLLM instance used for standard
rollouts, amortizing some of this cost. The overhead is proportional to
the number of cliff prompts per step.

A natural extension is to re-inject previously cliff prompts --- now
solvable after distillation --- back into the RL training distribution
for mode-sharpening. Crucially, re-injection should be delayed: by
allowing continued training on other prompts before revisiting former
cliffs, we test whether the distilled strategies are durably encoded
rather than transiently accessible. This expand-then-sharpen cycle could
systematically convert cliffs into solved prompts while maintaining the
benefits of RL's mode-seeking dynamics, achieving both high pass@k
(broad coverage) and high pass@1 (sharp modes). We leave this
curriculum-learning extension to future work.

\section{Conclusion}
\label{sec:conclusion}\label{conclusion}

We have introduced HDPO, a hybrid training objective that augments
reinforcement learning with privileged self-distillation to address the
cliff problem in mathematical reasoning. By leveraging ground truth as
privileged information and the model's own weights as the teacher, HDPO
provides bounded, non-zero gradients on prompts where the standard RL
gradient vanishes. Our theoretical analysis shows that same-model
privileged distillation achieves a strictly tighter realizability gap
than cross-model distillation, with the gap depending only on the
model's Lipschitz constant and the information content of the ground
truth (Proposition 1), and that R=1 filtering recovers the optimal
KL-regularized RL policy (Proposition 2).

Experiments on OpenMathInstruct-2 with Qwen2.5-Math-1.5B-Instruct show
that HDPO consistently improves \(pass@4\) and \(pass@8\), with the distillation weight \(\lambda\)
providing direct control over the exploration--exploitation tradeoff.
At $\lambda{=}0.1$, coverage improvements (pass@8 $+1.4$--$1.7\%$) reproduce robustly across hardware; at $\lambda{=}0.01$, improvements are consistent in direction but smaller in magnitude and more sensitive to run-to-run variance.

Looking forward, the expand-then-sharpen paradigm --- using HDPO to
broaden coverage, then RL to sharpen modes on previously unsolvable
prompts --- suggests a curriculum that could progressively reduce the fraction
of cliff prompts. Combined with evaluation on larger model scales
and diverse reasoning benchmarks, this direction could establish
privileged self-distillation as a standard component of RL training for
language models.

\appendix

\section{Realizability Gap Bound --- Full
Proof}

\subsection{Setup}\label{setup}

Let \(f_{\theta}\) denote a language model with parameters \(\theta\)
and vocabulary \(\mathcal{V}\). For context \(c\) (a sequence of token
embeddings), the model produces logits
\(z = f_{\theta}(c) \in \mathbb{R}^{\mathcal{|V|}}\) and output
distribution \(P_{\theta}( \cdot |c) = softmax(f_{\theta}(c))\).

For prompt \(x\), ground truth \(g\), and prefix \(y_{< t}\), define:

\begin{itemize}

\item
  \textbf{Teacher}: \(P_{T} = P_{\theta}( \cdot \mid c_{T})\) where
  \(c_{T} = \lbrack x;\, g;\, y_{< t}\rbrack\)
\item
  \textbf{Student}: \(P_{S} = P_{\theta}( \cdot \mid c_{S})\) where
  \(c_{S} = \lbrack x;\, y_{< t}\rbrack\)
\end{itemize}

The \textbf{realizability gap} at position \(t\) is
\(D_{KL}(P_{T} \parallel P_{S})\): the KL divergence between the teacher
and student distributions. A small realizability gap means the student
can feasibly match the teacher's distribution without large parameter
updates --- the optimization target is within reach.

\subsection{Lemma 1 (KL Divergence under Logit
Perturbation)}

This is a standard bound in information theory; we include the proof for
completeness.

\textbf{Statement.} \emph{For any} \(z \in \mathbb{R}^{\mathcal{|V|}}\)
\emph{and} \(\delta \in \mathbb{R}^{\mathcal{|V|}}\)\emph{:}

\[D_{KL}(softmax(z) \parallel softmax(z + \delta)) \leq \frac{\parallel \delta \parallel_{\infty}^{2}}{2}\]

\textbf{Proof.} Let \(P = softmax(z)\) and \(Q = softmax(z + \delta)\).
Write out the KL divergence explicitly:

\[D_{KL}(P \parallel Q) = \sum_{v\mathcal{\in V}}^{}P(v)log\frac{P(v)}{Q(v)}\]

Substituting \(P(v) = \frac{\exp\left( z_{v} \right)}{Z}\) and
\(Q(v) = \frac{\exp\left( z_{v} + \delta_{v} \right)}{Z'}\) where
\(Z = \sum_{j}^{}\exp(z_{j})\) and
\(Z' = \sum_{j}^{}\exp(z_{j} + \delta_{j})\):

\[D_{KL}(P \parallel Q) = \sum_{v\mathcal{\in V}}^{}P(v)\left( \log{P(v)} - \log{Q(v)} \right)\]

\[D_{KL}(P \parallel Q) = \sum_{v}^{}P(v)\left\lbrack z_{v} - logZ - z_{v} - \delta_{v} + logZ' \right\rbrack = - \mathbb{E}_{P}\lbrack\delta\rbrack + logZ' - logZ\]

\[\frac{Z'}{Z} = \frac{\sum_{j}^{}{e^{z_{j}}e^{\delta_{j}}}}{\sum_{j}^{}e^{z_{j}}} = \sum_{j}^{}{\frac{e^{z_{j}}}{\sum_{j}^{}e^{z_{j}}}e^{\delta_{j}}} = \sum_{j}^{}P_{j}exp(\delta_{j})\]

Since
\(\frac{Z'}{Z} = \sum_{j}^{}P(j)exp(\delta_{j}) = \mathbb{E}_{P}\lbrack exp(\delta)\rbrack\),
we obtain:

\[D_{KL}(P \parallel Q) = log\mathbb{E}_{P}\lbrack exp(\delta)\rbrack - \mathbb{E}_{P}\lbrack\delta\rbrack\]

This is the centered cumulant generating function of \(\delta\) under
\(P\).

Define
\(\psi(s) = log\mathbb{E}_{P}\lbrack exp(s\delta)\rbrack - s \cdot \mathbb{E}_{P}\lbrack\delta\rbrack\)
for \(s \in \lbrack 0,1\rbrack\). Then \(\psi(0) = 0\) and
\(\psi(1) = D_{KL}(P \parallel Q)\).

Computing derivatives:

\begin{itemize}

\item
  \(\psi'(s) = \mathbb{E}_{Q_{s}}\lbrack\delta\rbrack - \mathbb{E}_{P}\lbrack\delta\rbrack\),
  where \(Q_{s}(v) \propto P(v)exp(s \cdot \delta_{v})\) is the
  exponentially tilted distribution.
\item
  \(\psi''(s) = {Var}_{Q_{s}}(\delta)\)
\end{itemize}

By Taylor's theorem with integral remainder:

\[\psi(1) = \int_{0}^{1}(1 - s)\,\psi''(s)\, ds = \int_{0}^{1}(1 - s)\,{Var}_{Q_{s}}(\delta)\, ds\]

For any distribution over a bounded domain,
\(Var(X\mathbb{) \leq E\lbrack}X^{2}\rbrack \leq \parallel X \parallel_{\infty}^{2}\).
Since each component \(\delta_{v}\) satisfies
\(|\delta_{v}| \leq \parallel \delta \parallel_{\infty}\):

\[D_{KL}(P \parallel Q) \leq \parallel \delta \parallel_{\infty}^{2}\int_{0}^{1}(1 - s)\, ds = \frac{\parallel \delta \parallel_{\infty}^{2}}{2}\quad\quad\blacksquare\]

\subsection{Assumption 1 (Local Lipschitz
Continuity)}

On the domain of bounded inputs (finite embeddings, finite sequence
length), the logit function \(f_{\theta}\) is locally Lipschitz
continuous: there exists \(L_{\theta} < \infty\) such that

\[\parallel f_{\theta}(c_{1}) - f_{\theta}(c_{2}) \parallel_{\infty} \leq L_{\theta} \cdot d(c_{1},c_{2})\]

for all valid contexts \(c_{1},c_{2}\) in the bounded domain, where
\(d( \cdot , \cdot )\) is a distance metric on the input space.

\textbf{Justification.} This is a standard property for neural networks
on compact domains. For transformers specifically, \citet{Kim2021}
establish local Lipschitz bounds for self-attention on bounded input
domains. Recent work shows the per-layer local Lipschitz constant scales
as \(O(C\sqrt{n})\) where \(n\) is sequence length and \(C\) depends on
weight norms and attention distribution concentration.

\subsection{Proposition 1 (Realizability Gap
Comparison)}

\subsubsection{Part I -- Same-Model
Bound}

\textbf{Statement.} \emph{Under Assumption 1, the per-position
realizability gap for same-model privileged distillation satisfies:}

\[D_{KL}(P_{T} \parallel P_{S}) \leq \frac{L_{\theta}^{2} \cdot \Delta(g)^{2}}{2}\]

\emph{where} \(\Delta(g) = d(c_{T},c_{S})\) \emph{is the input-space
distance attributable to the ground truth tokens} \(g\)\emph{.}

\textbf{Proof.} Teacher and student evaluate the same function
\(f_{\theta}\) on different inputs \(c_{T}\) and \(c_{S}\). By
Assumption 1:

\[\parallel z_{T} - z_{S} \parallel_{\infty} = \parallel f_{\theta}(c_{T}) - f_{\theta}(c_{S}) \parallel_{\infty} \leq L_{\theta} \cdot d(c_{T},c_{S}) = L_{\theta} \cdot \Delta(g)\]

Applying Lemma 1 with \(\delta = z_{T} - z_{S}\):

\[D_{KL}(P_{T} \parallel P_{S}) \leq \frac{\parallel z_{T} - z_{S} \parallel_{\infty}^{2}}{2} \leq \frac{L_{\theta}^{2} \cdot \Delta(g)^{2}}{2}\quad\quad\blacksquare\]

\textbf{Key properties of this bound:}

\begin{enumerate}
\def\labelenumi{\arabic{enumi}.}

\item
  It depends only on the model \(\theta\) (via \(L_{\theta}\)) and the
  information content of the ground truth \(g\) (via \(\Delta(g)\)).
\item
  It does not depend on any capacity gap between two different models.
\item
  The difficulty of distillation scales with how much the ground truth
  changes the prediction, not with architectural differences.
\end{enumerate}

\subsubsection{Part II -- Cross-Model
Comparison}

\textbf{Statement.} \emph{Under Assumption 1, for cross-model
distillation with teacher} \(\phi\) \emph{and student} \(\theta\)\emph{,
where the teacher receives privileged input} \(c_{T}\) \emph{and the
student receives} \(c_{S}\)\emph{, the realizability gap satisfies:}

\[D_{KL}(P_{\phi}( \cdot |c_{T}) \parallel P_{\theta}( \cdot |c_{S})) \leq \frac{\left( L_{\theta} \cdot \Delta(g) + \left\| f_{\phi}\left( c_{T} \right) - f_{\theta}\left( c_{T} \right) \right\|_{\infty} \right)^{2}}{2}\]

\textbf{Proof.} The logit difference between the cross-model teacher and
the student decomposes as:

\[f_{\phi}\left( c_{T} \right) - f_{\theta}\left( c_{S} \right) = \left\lbrack f_{\phi}\left( c_{T} \right) - f_{\theta}\left( c_{T} \right) \right\rbrack + \left\lbrack f_{\theta}\left( c_{T} \right) - f_{\theta}\left( c_{S} \right) \right\rbrack\]

The first bracket captures the model mismatch: the difference in logit
functions between models \(\phi\) and \(\theta\) evaluated on the same
input. The second bracket is the same input perturbation that appears in
Part I. By the triangle inequality:

\[\left\| f_{\phi}\left( c_{T} \right) - f_{\theta}\left( c_{S} \right) \right\|_{\infty} \leq \left\| f_{\phi}\left( c_{T} \right) - f_{\theta}\left( c_{T} \right) \right\|_{\infty} + L_{\theta} \cdot \Delta(g)\]

where the second inequality applies Assumption 1 to the student model
\(\theta\). Applying Lemma 1 with
\(\delta = f_{\phi}\left( c_{T} \right) - f_{\theta}\left( c_{S} \right)\)
yields the bound. $\blacksquare$

\subsection{Remark}\label{remark}

The bound in Part I may be quantitatively loose: the Lipschitz constant
\(L_{\theta}\) for a deep transformer can be large due to multiplicative
composition across layers. However, the comparison in Part II does not
depend on the tightness of the Lipschitz bound. The same
\(L_{\theta} \cdot \Delta(g)\) term appears in both bounds, and the
cross-model bound is strictly larger by the model-mismatch term
regardless of the absolute magnitudes. A loose Lipschitz constant
affects both bounds equally --- it does not favor one distillation
setting over the other.

Moreover, \(L_{\theta}\) is the same for both teacher and student in the
same-model case (since they share parameters), so it does not introduce
a capacity mismatch --- it merely scales the bound uniformly.

\section{Proposition 2 Full
Proof}

\textbf{Proposition 2} (\(R = 1\) Filtering Yields the RL-Optimal
Policy). We provide the full derivation here. For binary reward:

\emph{For binary reward} \(R(\tau)\  \in \ \{ 0,\ 1\}\)\emph{, the
exponential factor takes only two values:}

\[\exp\left( \frac{R(\tau)}{\beta} \right)\  = \ exp\left( \frac{1}{\beta} \right)\ \ if\ R(\tau)\  = \ 1,\ \ \ \ \ and\ \ \ \ \ exp\left( \frac{R(\tau)}{\beta} \right)\  = \ 1\ \ if\ R(\tau)\  = \ 0\]

\emph{Substituting into the Gibbs distribution:}

\[\pi^{*}(\tau)\  = \ \frac{\pi_{ref}(\tau)\  \cdot \ exp\left( \frac{1\left\lbrack R(\tau) = \ 1 \right\rbrack}{\beta} \right)}{Z(\beta)}\]

\emph{where the partition function evaluates to:}

\[Z(\beta)\  = \ P_{\pi_{ref}}(R\  = \ 0)\  \cdot \ 1\  + \ P_{\pi_{ref}}(R\  = \ 1)\  \cdot \ exp\left( \frac{1}{\beta} \right)\  = \ (1\  - \ p)\  + \ p\  \cdot \ exp\left( \frac{1}{\beta} \right)\]

\emph{where} \(p\  = \ P_{\pi\_ ref}(R\  = \ 1)\) \emph{is the
probability that} \(\pi_{ref}\) \emph{generates a correct trajectory.
This gives the relative weighting of correct vs incorrect trajectories:}

\[\frac{\pi^{*}(\tau\ |\ R\  = \ 1)}{\pi^{*}(\tau\ |\ R\  = \ 0)}\  = \ exp\left( \frac{1}{\beta} \right)\ \ \ \ \ (\text{for trajectories with equal }\pi_{\text{ref}}\text{ weight})\]

\emph{\textbf{The hard-threshold limit.} As}
\(\beta\  \rightarrow \ 0^{+}\)\emph{, the exponential ratio}
\(exp(1/\beta)\  \rightarrow \ \infty\)\emph{. This means correct
trajectories receive infinitely more weight than incorrect ones.
Concretely, for any incorrect trajectory with}
\(R(\tau)\  = \ 0\)\emph{:}

\[\pi^{*}(\tau)\  = \ \frac{\pi_{ref}(\tau)\  \cdot \ exp\left( \frac{1\lbrack 0 = \ 1\rbrack}{\beta} \right)}{Z(\beta)}\  = \ \frac{\pi_{ref}(\tau)\  \cdot \ 1}{Z(\beta)}\  \rightarrow \ 0\ \ \ \ \ as\ \beta\  \rightarrow \ 0^{+}\]

\emph{because} \(Z(\beta)\  \sim \ p\  \cdot \ exp(1/\beta)\)
\emph{grows without bound while the numerator is fixed. Conversely, for
a correct trajectory with} \(R(\tau)\  = \ 1\)\emph{:}

\[\pi^{*}(\tau)\  = \ \frac{\pi_{ref}(\tau)\  \cdot \ exp\left( \frac{1\lbrack 1\  = \ 1\rbrack}{\beta} \right)}{Z(\beta)}\  = \ \frac{\pi_{ref}(\tau)\  \cdot \ exp\left( \frac{1}{\beta} \right)}{p\  \cdot \ exp\left( \frac{1}{\beta} \right)\  + \ (1\  - \ p)}\  \rightarrow \ \frac{\pi_{ref}(\tau)}{p}\  = \ \frac{\pi_{ref}(\tau)}{P_{\pi\_ ref}(R\  = \ 1)}\]

\emph{where the limit follows from dividing numerator and denominator
by} \(exp(1/\beta)\) \emph{and noting that}
\((1\  - \ p)\ /\ exp(1/\beta)\  \rightarrow \ 0\)\emph{. Therefore, in
the limit:}

\[\pi^{*}(\tau)\  \rightarrow \ \pi_{ref}(\tau\ |\ R(\tau)\  = \ 1)\  = \ \frac{\pi_{ref}(\tau)\  \cdot \ 1\lbrack R(\tau)\  = \ 1\rbrack}{P_{\pi\_ ref}(R\  = \ 1)}\]

\section{Experimental
Details}

\begin{table}[t]
\centering
\caption{Hyperparameters used in all experiments.}
\label{tab:hyperparams}
\small
\begin{tabular}{lll}
\toprule
\textbf{Category} & \textbf{Parameter} & \textbf{Value} \\
\midrule
Model & Architecture & Qwen2.5-Math-1.5B-Instruct \\
      & Precision & bfloat16 \\
      & Max sequence length & 4096 \\
Dataset & Training & OpenMathInstruct-2 \\
        & Validation samples & 2048 \\
        & Reward function & hf\_math\_verify (binary) \\
Optimization & Optimizer & AdamW ($\beta_1{=}0.9$, $\beta_2{=}0.999$, $\epsilon{=}10^{-8}$) \\
             & Learning rate & $10^{-6}$ \\
             & LR schedule & Linear warmup (50 steps, 0.1$\to$1.0$\times$), then constant \\
             & Weight decay & 0.01 \\
             & Max gradient norm & 1.0 \\
GRPO & Prompts per step & 32 \\
     & Generations per prompt & 16 \\
     & Training batch size & 512 \\
     & Advantage normalization & Leave-one-out baseline \\
     & Ratio clip $\epsilon$ & 0.2 \\
     & Loss granularity & Token-level \\
     & Reference KL penalty $\beta$ & 0.0 \\
     & Generation temperature & 1.0 \\
     & Generation backend & vLLM (colocated) \\
     & Sequence packing & Enabled \\
Privileged Distillation & Distillation weight $\lambda$ & \{0.01, 0.1\} \\
                        & Distillation loss & JSD \\
                        & Teacher type & \{Drifting (shared current policy weights), Frozen (initial weights)\} \\
                        & Normalization & Global token count (rank-invariant) \\
                        & Teacher top-$k$ & 64 \\
                        & Cliff threshold & reward\_sum $= 0.0$ \\
                        & Teacher success threshold & reward $\geq 1.0$ \\
                        & Max cliff prompts per step & 32 \\
Evaluation & Validation period & Every 10 steps \\
           & Reported metrics & pass@1, pass@4, pass@8 \\
Infrastructure & GPUs & 8$\times$H200 (1 node); replicated on 8$\times$H100 \\
               & Seed & 42 \\
\bottomrule
\end{tabular}
\end{table}

\section{Hardware Variation}
\label{app:hardware}

All main results (Table~\ref{tab:results}) were obtained on 8$\times$H200 GPUs. We replicated all five configurations on 8$\times$H100 GPUs with identical hyperparameters. The training computation is mathematically equivalent; differences arise from floating-point non-determinism across GPU microarchitectures (different matmul tiling, reduction order, and FlashAttention kernel implementations), which compound over 2000 training steps to produce subtly different final weights.

\begin{table}[h]
\centering
\caption{Best pass@$k$ on 8$\times$H100 GPUs. Same setup as Table~\ref{tab:results}.
Bold indicates best in column.}
\label{tab:results-h100}
\vspace{2pt}
\begin{tabular}{lccc}
\toprule
\textbf{Method} & \textbf{pass@1} & \textbf{pass@4} & \textbf{pass@8} \\
\midrule
GRPO Baseline                & \textbf{0.6509} & 0.7739          & 0.8223 \\
HDPO (frozen, $\lambda$=0.01)   & 0.6484          & 0.7773          & 0.8252 \\
HDPO (frozen, $\lambda$=0.1)    & 0.6343          & \textbf{0.7856} & \textbf{0.8369} \\
HDPO (drifting, $\lambda$=0.01) & 0.6499          & 0.7783          & 0.8213 \\
HDPO (drifting, $\lambda$=0.1)  & 0.6343          & 0.7832          & 0.8359 \\
\bottomrule
\end{tabular}
\end{table}

The qualitative findings are consistent: $\lambda{=}0.1$ improves pass@8 by $+1.4$--$1.5\%$ over baseline on both hardware configurations, and pass@1 remains highest for the baseline and $\lambda{=}0.01$ configurations. The primary quantitative difference is at $\lambda{=}0.01$: on H200, drifting-$0.01$ achieves the highest pass@4 ($+1.1\%$), while on H100 the improvement is smaller ($+0.4\%$) and the best pass@4 shifts to frozen-$0.1$. This suggests the $\lambda{=}0.01$ improvements, while directionally consistent, are close to the noise floor introduced by hardware-level floating-point non-determinism and temperature-1 evaluation variance. The $\lambda{=}0.1$ coverage improvements are robust across both configurations.

\section*{LLM Usage Statement}

The research idea underlying HDPO---using privileged self-distillation to provide
learning signal on cliff prompts---originated with the author. However, an AI
language model (Claude, Anthropic) was used extensively throughout this project in
ways that go beyond minor writing assistance. Specifically: (1)~the mathematical
formalization of Proposition~1 and Proposition~2, including the proof structure and
verification of correctness, was developed collaboratively with LLM assistance;
(2)~the paper text, including the related work survey, theoretical exposition, and
discussion sections, was substantially drafted and edited with LLM assistance;
and (3)~the LLM was used as a research collaborator to brainstorm explanations
for experimental phenomena (e.g., the pass@1 vs.\ pass@$k$ tradeoff) and to
compare HDPO against related work.
All experimental results (training runs, metric measurements) were produced
by the author without LLM involvement. The LLM also provided substantial
code implementation assistance for the training infrastructure.

\bibliography{hdpo_arxiv}
\bibliographystyle{plainnat}

\end{document}